\documentclass[sigconf]{acmart}
\usepackage{amsmath}
\usepackage{algorithm}
\usepackage{algpseudocode}
\usepackage[normalem]{ulem}
\usepackage[utf8]{inputenc}
\setcitestyle{numbers,sort&compress}

\AtBeginDocument{%
  \providecommand\BibTeX{{\normalfont B\kern-0.5em{\scshape i\kern-0.25em b}\kern-0.8em\TeX}}%
}

\renewcommand\footnotetextcopyrightpermission[1]{}
\makeatletter
\def\@ACM@title@font{\Large\bfseries}
\def\@ACM@subtitlefont{\normalsize\itshape}
\makeatother
\acmConference[DSE 2026]{International Workshop on Dark Software Engineering (DSE 2026)}{April 12--18, 2026}{Rio de Janeiro, Brazil}
\acmBooktitle{Proceedings of the 48th International Conference on Software Engineering (ICSE 2026), April 12--18, 2026, Rio de Janeiro, Brazil}

\begin{document}

\title{The Moral Consistency Pipeline: Continuous Ethical Evaluation for Large Language Models}

\author{Saeid Jamshidi, Kawser Wazed Nafi, Arghavan Moradi Dakhel, Negar Shahabi$^{\dagger}$, Foutse Khomh}
\affiliation{%
  \institution{SWAT Laboratory, Polytechnique Montréal}
  \institution{$^{\dagger}$Concordia Institute for Information Systems Engineering, Concordia University}
  \city{Montréal}
  \state{Quebec}
  \country{Canada}
}
\email{{saeid.jamshidi, kawser.wazed-nafi, arghavan.moradi-dakhel, foutse.khomh}@polymtl.ca, negar.shahabi@concordia.ca}

  

\begin{abstract}
The rapid advancement and adaptability of Large Language Models (LLMs) highlight the need for \textit{moral consistency}, the capacity to maintain ethically coherent reasoning across varied contexts. Existing \textit{alignment frameworks}, structured approaches designed to align model behavior with human ethical and social norms, often rely on static datasets and post-hoc evaluations, offering limited insight into how ethical reasoning may evolve across different contexts or temporal scales. This study presents the \textit{Moral Consistency Pipeline (MoCoP)}, a dataset-free, closed-loop framework for continuously evaluating and interpreting the moral stability of LLMs. MoCoP combines three supporting layers: (i) lexical integrity analysis, (ii) semantic risk estimation, and (iii) reasoning-based judgment modeling within a self-sustaining architecture that autonomously generates, evaluates, and refines ethical scenarios without external supervision. Our empirical results on GPT-4-Turbo and DeepSeek suggest that MoCoP effectively captures longitudinal ethical behavior, revealing a strong inverse relationship between ethical and toxicity dimensions ($r_{ET}=-0.81$, $p<0.001$) and a near-zero association with response latency ($r_{EL}\approx0$). These findings demonstrate that moral coherence and linguistic safety tend to emerge as stable and interpretable characteristics of model behavior rather than short-term fluctuations. Furthermore, by reframing ethical evaluation as a dynamic, model-agnostic form of moral introspection, MoCoP offers a reproducible foundation for scalable, continuous auditing and advances the study of computational morality in autonomous AI systems.
\end{abstract}

\maketitle
\section{Introduction}
\label{Intro}
Artificial intelligence (AI) systems are rapidly being integrated into decision-making, communication, and learning environments, where their growing autonomy makes ethical awareness, moral coherence, and interpretability increasingly essential~\cite{khreisat2024ethical, nanjundan2025navigating, bayan2024ethics}. As Large Language Models (LLMs) evolve from linguistic tools to autonomous agents, capable of reasoning and adaptation, they begin to impact human judgment, social perception, and institutional processes~\cite{huang2024levels, fang2025comprehensive, barua2024exploring}. This transformation underscores the need for frameworks that examine not only what these models produce, but also how and why their moral reasoning remains coherent across different contexts over time~\cite{akinrinola2024navigating, mcintosh2025inadequacies}.\\
Existing safety-alignment and interpretability frameworks have made progress in bias detection, toxicity reduction, and fairness auditing~\cite{shi2024large, lu2025alignment}. However, these frameworks generally evaluate isolated model outputs rather than tracing the evolution of ethical reasoning or shifts in value alignment over time as models interact with diverse, context-dependent, and dynamically evolving scenarios~\cite{mclaren2006computational}. This gap is particularly significant in the context of LLM-based agentic systems, which exhibit adaptive and context-sensitive behavior, continuously modifying their outputs in response to prompt structure, user intent, task requirements, and prior interactions. The absence of such longitudinal consistency raises the risk of \textit{moral drift}—a phenomenon in which ethical stances fluctuate and degrade in response to prompt wording, contextual pressures, and temporal changes. Inconsistencies of this kind can threaten the reliability, accountability, and societal trustworthiness of contemporary AI systems~\cite{bina2023agency, tenbrunsel2010ethical}.\\
Ensuring moral consistency is therefore not only a matter of regulatory compliance but also a prerequisite for maintaining logically stable, interpretable, and trustworthy reasoning within LLM-driven AI systems over time~\cite{agrawal2024accountability, zhang2023ethics}. While developing autonomous systems in high-stakes domains such as healthcare, finance, education, and governance, even a small ethical instability can propagate misinformation, reinforce cognitive bias, and produce outcomes misaligned with human values. Despite growing recognition of these challenges, most current auditing techniques, used as part of broader safety-alignment frameworks, remain constrained by static benchmarks and fixed, human-annotated datasets that fail to adapt to the evolving and context-dependent behavior of LLMs. Consequently, a key research challenge is to design an adaptive, self-contained mechanism capable of autonomously evaluating and interpreting the ethical stability of such models without dependence on external datasets or post-hoc analysis.\\
To address this challenge, this work introduces the \textit{Moral Consistency Pipeline (MoCoP)}, a fully autonomous, dataset-free ethical evaluation framework designed to continuously measure, interpret, and refine the moral reasoning patterns of LLMs. MoCoP operates as a closed-loop experimental system that dynamically generates ethical scenarios, elicits model responses, and performs structured interpretation using internally defined rules and semantic heuristics. In contrast to traditional evaluation approaches that depend on predefined test corpora and external annotation, MoCoP autonomously constructs and analyzes its own ethical scenarios. It integrates three complementary analytical components—\textit{lexical integrity analysis}, \textit{semantic risk estimation}, and \textit{reasoning-based judgment modeling}—to derive unified measures of moral coherence and ethical stability. Each layer contributes a distinct function: lexical integrity detects bias and linguistic degradation, semantic risk estimation quantifies contextual harm such as toxicity or coercion, and reasoning-based judgment modeling evaluates the logical soundness and ethical structure of responses. Through iterative self-analysis and adaptive evaluation, MoCoP reframes ethical assessment as a proactive, self-sustaining process that evolves in tandem with the models it studies.\\
The MoCoP framework combines three core subsystems that operationalize these analytical layers: (1) an \textbf{LLMConnector}, which enables controlled and reproducible interactions between heterogeneous models such as GPT-4-Turbo\footnote{OpenAI GPT-4-Turbo model, 2024} and DeepSeek; (2) an \textbf{EthicalGuardPro} engine, which performs lexical scanning, semantic filtering, and moral interpretation through rule-based and probabilistic reasoning; and (3) a \textbf{Meta-Analytic Ethics Layer}, which aggregates data from multiple experiments to compute divergence, coherence, and consensus metrics across large-scale evaluations. Together, these components facilitate autonomous cross-model benchmarking, dynamic tracking of moral evolution, and quantitative performance analysis, providing the foundation for longitudinal ethical introspection in LLMs. The main contributions of this paper are summarized as follows:
\begin{itemize}
    \item \textbf{Autonomous Moral Evaluation Framework:} Introduction of MoCoP, a dataset-free, self-sustaining system that autonomously generates, evaluates, and refines moral reasoning scenarios across multiple LLMs without external supervision.
    \item \textbf{Cross-Model Moral Drift Quantification:} Development of quantitative methods for measuring moral drift, ethical stability, and inter-model coherence, enabling comparative analysis between distinct LLM architectures.
    \item \textbf{Unified Meta-Analytic Ethics Engine:} Integration of lexical, semantic, and reasoning-based evaluation modules into a reproducible framework for continuous ethical introspection, providing longitudinal insight into the stability and evolution of LLM moral behavior.
\end{itemize}
The remaining sections of this paper are organized as follows. Section~\ref{Related Work} reviews related research on \textit{evaluating the ethical dimensions of outputs generated by LLMs}. Section~\ref{sec:methodology} details the proposed MoCoP framework, followed by experimental results in Section~\ref{Experimental Results}. Section~\ref{Discussion} discusses findings, Section~\ref{Threats to Validity} outlines threats to validity, and Section~\ref{Conclusion, Limitations, and Future Work} concludes with limitations and future directions.

\section{Related Work}
\label{Related Work}
Early research focused primarily on detecting and mitigating social bias and harmful content generation. Burak et al. \cite{ccetin2025openethics} introduced \textit{OpenEthics}, emphasizing transparency and open evaluation pipelines for auditing ethical alignment in LLMs. Similarly, M Abdulhai et al. \cite{abdulhai2023moral} explored rule-based fairness evaluation and post-hoc interpretability to reduce discriminatory model behavior across demographic groups. Although these studies advanced auditing transparency, their reliance on fixed datasets and manually curated prompts limits adaptability. They provide diagnostic insights into static model states but cannot continuously track shifts in moral reasoning and measure stability across varying contexts.\\
Recent approaches have extended beyond fairness to explore moral judgment and ethical reasoning capacities in LLMs. Studies such as \cite{ungless2024only} and \cite{ji2025moralbench} proposed structured moral reasoning benchmarks (e.g., \textit{MoralBench}) to test LLM responses against human moral intuitions. These works demonstrated that models exhibit partial moral sensitivity but often lack coherence when faced with conflicting principles and contextual ambiguity. However, benchmark-based evaluations remain limited in scalability and temporal scope; they capture model ethics at a single point in time and cannot reveal how reasoning consistency evolves with interaction and retraining.\\
Supporting studies have examined explainability and value alignment as central components of responsible AI. For example, Shitong et al. \cite{duan2023denevil} proposed \textit{DeNEVIL}, a debiasing and explainable alignment model that integrates normative reasoning into LLM training. Likewise, Jiao et al. \cite{jiao2025llm} emphasized cross-cultural moral benchmarking and trust calibration, underscoring the sociotechnical dependencies of ethical AI systems. Despite their strengths in interpretability and trust analysis, these frameworks rely on predefined moral categories and manual labeling, which limit their adaptability to dynamic ethical contexts. \\
Although previous literature collectively contributes to fairness auditing, benchmark design, and value-alignment methodologies, a key gap remains unaddressed: the absence of a dynamic, model-agnostic framework capable of longitudinal moral evaluation. Existing works assess ethical behavior at discrete points, but they lack mechanisms for autonomous generation, interpretation, and adaptation of ethical reasoning scenarios. Moreover, none of the existing approaches provides computational tools to quantify \textit{ethical divergence} or track \textit{cross-model moral coherence} in real time.\\


\section{Methodology}
\label{sec:methodology}
Building upon the conceptual foundation established in Section~\ref{Intro}, the proposed \textit{Moral Consistency Pipeline (MoCoP)} operationalizes continuous ethical evaluation as a closed-loop, dataset-free mechanism designed to measure, interpret, and refine the moral stability of Large Language Models (LLMs). In alignment with the abstract and introduction, MoCoP focuses exclusively on \textbf{evaluation and interpretation}, rather than optimization or retraining, ensuring conceptual and methodological coherence. 

\subsection{System Overview}
\begin{figure}[h!]
    \centering
    \includegraphics[width=0.90\linewidth]{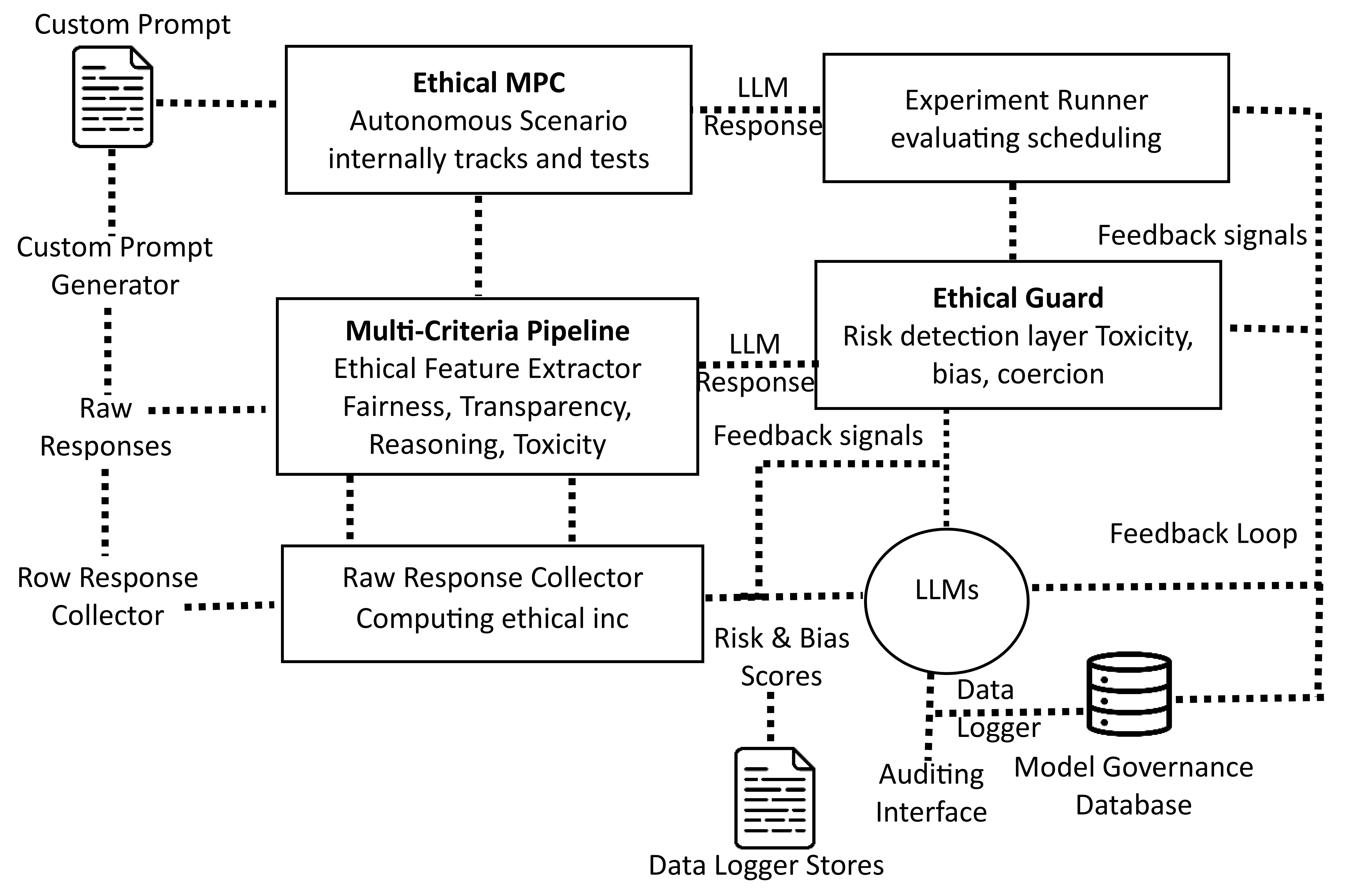}
    \caption{Schematic of the Architecture of the MoCoP Framework. The pipeline integrates lexical integrity analysis, semantic risk estimation, and reasoning-based judgment modeling in a continuous feedback loop for evaluating LLM moral consistency. Arrows indicate feedback flow across layers.}
    \label{fig:ethical_mcp_architecture}
\end{figure}

As illustrated in Figure~\ref{fig:ethical_mcp_architecture}, MoCoP follows a feedback-driven architecture composed of three primary analytical layers: (i) \textit{Lexical Integrity Analysis}, (ii) \textit{Semantic Risk Estimation}, and (iii) \textit{Reasoning-Based Judgment Modeling}. Together, these layers quantify moral consistency across contexts and over time. This proposed architecture autonomously generates ethically sensitive scenarios, evaluates model responses, and iteratively refines scenario complexity using internal feedback mechanisms—without reproducing or amplifying unethical content.

\subsection{Core Analytical Layers} \label{metricsanalysis}
\paragraph{Lexical Integrity Analysis.} 
This layer evaluates surface-level linguistic structure to detect bias, fairness deviations, and polarity inconsistencies. Using lexical entropy, polarity variance, and bias-weighted sentiment, the system estimates how consistently models maintain neutral and coherent language across ethical contexts. This produces the lexical integrity feature $L_{ij}$ for each model response $r_{ij}$.

\paragraph{Semantic Risk Estimation.}
The second layer assesses contextual risk through probabilistic semantic embedding analysis. It quantifies potential harm, coercion, and toxicity using hybrid lexical–semantic similarity and bounded risk functions. The resulting score, $\tau_{ij}$, represents each model’s estimated semantic toxicity index, encoding its ethical exposure to risk.

\paragraph{Reasoning-Based Judgment Modeling.}
This layer captures logical and ethical soundness within model reasoning. Each response is decomposed into propositional reasoning chains that are evaluated for moral justification, causal coherence, and logical stability. The derived feature $R_{ij}$ represents the strength of ethical reasoning consistency. Together, the ethical feature vector for each model–prompt pair $(M_j, p_i)$ is expressed as:
\[
\mathbf{E}_{ij} = [L_{ij}, \tau_{ij}, R_{ij}],
\]
forming the foundation of all subsequent quantitative analyses.

\subsection{Formal Framework}
Let $\mathcal{P} = \{p_1, p_2, \dots, p_N\}$ denote the set of automatically generated moral scenarios, and $\mathcal{M} = \{M_1, M_2\}$ represent the evaluated LLMs—GPT-4-Turbo and DeepSeek. For each prompt–model pair, the response $r_{ij} = M_j(p_i)$ is projected into the ethical feature space via:
\[
\mathbf{E}_{ij} = \text{Eval}(r_{ij}) = [L_{ij}, \tau_{ij}, R_{ij}].
\]
An overall \textit{Ethical Utility Function} measures combined moral performance:
\[
J(\theta) = \mathbb{E}_{i,j}\,[\, \alpha L_{ij} + \beta R_{ij} - \lambda \tau_{ij} \,],
\]
where $\alpha$, $\beta$, and $\lambda$ are adaptive weights controlling the trade-off between lexical coherence, reasoning alignment, and toxicity reduction. The system seeks stability when $J(\theta)$ converges toward a steady state across evaluation cycles.

\subsection{Feedback Dynamics}
Each evaluation cycle consists of the following sequence:
\begin{enumerate}
    \item \textbf{Prompt Generation:} Sample or regenerate a scenario $p_i$ from a moral domain distribution.
    \item \textbf{Response Collection:} Query each model $M_j$ and collect responses $r_{ij}$.
    \item \textbf{Feature Extraction:} Compute the ethical feature vector $\mathbf{E}_{ij}$.
    \item \textbf{Aggregation \& Adaptation:} Aggregate cross-model metrics, compute divergence, and update the prompt distribution to maintain balance.
\end{enumerate}
This self-sustaining loop ensures MoCoP continuously evaluates model behavior under evolving ethical conditions without human annotation or external supervision.

\begin{algorithm}[H]
\caption{MoCoP Evaluation Cycle}
\label{alg:mocop}
\begin{algorithmic}[1]
\Require Prompt set $\mathcal{P}$, model set $\mathcal{M}=\{M_1,M_2\}$, initial weights $\theta=(\alpha,\beta,\lambda)$, learning rate $\eta_\theta$
\Ensure Ethical utility trajectory $J(\theta)$ and convergence statistics
\For{each prompt $p_i \in \mathcal{P}$}
    \For{each model $M_j \in \mathcal{M}$}
        \State Generate response $r_{ij} \gets M_j(p_i)$
        \State Extract ethical features $\mathbf{E}_{ij} = [L_{ij}, \tau_{ij}, R_{ij}]$
        \State Compute ethical utility $J_{ij} = \alpha L_{ij} + \beta R_{ij} - \lambda \tau_{ij}$
    \EndFor
    \State Calculate inter-model divergence $\mathcal{D}_{moral}$
    \State Update scenario distribution using feedback regulator $\mathcal{F}$
\EndFor
\State Update parameters: $\theta \gets \theta - \eta_\theta \nabla_\theta J(\theta)$
\State Check convergence: $\Delta J^{(t)} < \varepsilon$
\end{algorithmic}
\end{algorithm}

Algorithm~\ref{alg:mocop} formalizes the MoCoP feedback loop described above. Each iteration autonomously generates scenarios, extracts ethical features, and aggregates model outputs to refine the distribution of ethical stimuli. Unlike static benchmark evaluations, MoCoP iteratively adjusts prompt difficulty and ethical context weighting, allowing dynamic assessment of moral stability while maintaining dataset independence.

\subsection{Quantitative Convergence}
Ethical equilibrium is achieved when the utility function variation stabilizes:
\[
\lim_{t \to \infty} \Delta J^{(t)} = 0,
\]
where $\Delta J^{(t)}$ measures inter-iteration change. Convergence indicates that model moral reasoning has reached consistent performance across both contextual and temporal dimensions, fulfilling MoCoP’s objective of longitudinal ethical consistency.

\section{Experimental Setup}
\label{sec:experimental_setup}
To empirically validate MoCoP’s capacity for continuous, interpretable ethical evaluation, we implemented the framework in a closed-loop experimental environment reflecting the three analytical layers defined in Section~\ref{sec:methodology}. The experimental objectives include assessing (i) cross-model moral coherence, (ii) temporal stability, and (iii) semantic safety across evolving ethical scenarios.

\subsection{Environment Configuration}
The MoCoP environment is defined as:
\[
\mathcal{MoCoP} = \langle \mathcal{I}, \mathcal{E}, \mathcal{M}, \mathcal{F} \rangle,
\]
where:
\begin{itemize}
    \item $\mathcal{I}$ (\textbf{Interaction Layer}) manages structured input–output communication and ensures prompt standardization.
    \item $\mathcal{E}$ (\textbf{Ethical Evaluation Engine}) implements the three-layer analysis (lexical, semantic, reasoning).
    \item $\mathcal{M}$ (\textbf{Meta-Analytic Layer}) aggregates results, computes divergence, and tracks moral drift.
    \item $\mathcal{F}$ (\textbf{Feedback Regulator}) adjusts prompt distributions to maintain balanced evaluation entropy and convergence.
\end{itemize}
Each prompt evolves over iterations as:
\[
p_j^{(t+1)} = \mathcal{F}\big(\mathcal{E}(\mathcal{I}(M_i, p_j^{(t)}))\big),
\]
producing a self-adapting testbed that maintains ethical continuity and variability across evaluation cycles.

\subsection{Models Under Evaluation}
Two representative LLMs were selected:
\begin{enumerate}
    \item \textbf{GPT-4-Turbo:} A transformer-based model emphasizing factual precision and safety-aligned moral reasoning.
    \item \textbf{DeepSeek:} A reinforcement-augmented generative model emphasizing interpretive depth and contextual adaptability.
\end{enumerate}
Both models were tested under identical conditions, ensuring that observed differences emerge from intrinsic reasoning patterns rather than external variance.

\subsection{Autonomous Moral Scenario Generation}
To preserve dataset independence, MoCoP autonomously generated $n=500$ unique ethical prompts spanning five domains:
\[
\mathcal{P} = \mathcal{P}_{fairness} \cup \mathcal{P}_{privacy} \cup \mathcal{P}_{transparency} \cup \mathcal{P}_{coercion} \cup \mathcal{P}_{alignment}.
\]
Each domain’s weight $w_c$ was dynamically updated according to divergence magnitude, ensuring balanced exploration. Prompts were bounded by lexical entropy $\mathcal{H}(p_j) < 0.7$ to prevent degeneration and maintain linguistic clarity.

\subsection{Evaluation Metrics}
For each model($M_{i}$)–prompt($p_{j}$) pair, MoCoP computes:
\[
S_{ij} = (s^{(lex)}_{ij}, s^{(sem)}_{ij}, s^{(rea)}_{ij}),
\]
Where $s^{(lex)}_{ij}$, $s^{(sem)}_{ij}$ and $s^{(rea)}_{ij}$ denote the Lexical Integrity \cite{han2024evaluation}, Semantic Safety \cite{radcliffe2024automated}, and Reasoning Coherence \cite{do2025defines} scores, respectively, as detailed in Section~\ref{metricsanalysis}. Each metric produces a normalized score within the range [0,1], with higher values indicating stronger adherence to safety and reliability criteria. Specifically, Lexical Integrity assesses the preservation of prompt structure and resistance to injection or tampering; Semantic Safety evaluates alignment with intended meaning and absence of harmful or biased content; and Reasoning Coherence measures the logical consistency and causal validity of the model’s reasoning chain. A score approaching 1.0 in any dimension signifies optimal performance for that specific safety attribute. Reasoning coherence is measured by cosine similarity:
\[
s^{(rea)}_{ij} = \cos(\phi(p_j), \phi(r_{ij})),
\]
while semantic safety is defined via a logistic toxicity function:
\[
s^{(sem)}_{ij} = 1 - \sigma(\alpha \cdot \text{tox}(r_{ij})), \quad \alpha = 0.85.
\]
The final composite ethical response is calculated as:
\[
\mathcal{R}(M_i, p_j) = w_1 s^{(lex)}_{ij} + w_2 s^{(sem)}_{ij} + w_3 s^{(rea)}_{ij},
\]
where $(w_1, w_2, w_3) = (0.3, 0.35, 0.35)$, emphasizing balance between linguistic neutrality and reasoning soundness.

\subsection{Performance and Stability Metrics}
Global ethical consistency is defined as:
\[
\text{ECI}(M_i) = \mathbb{E}_{p_j \sim \mathcal{P}}[\mathcal{R}(M_i, p_j)],
\]
while cross-model divergence and temporal stability are quantified as:
\[
\mathcal{D}_{moral} = \frac{1}{n}\sum_{j=1}^{n} \big| \text{ECI}_{GPT-4}^{(j)} - \text{ECI}_{DeepSeek}^{(j)} \big|,
\]
\[
\mathcal{S}_{temporal} = 1 - \frac{1}{T-1} \sum_{t=1}^{T-1} \big| \text{ECI}^{(t+1)} - \text{ECI}^{(t)} \big|.
\]
Higher $\mathcal{S}_{temporal}$ values reflect stronger moral steadiness and ethical robustness across feedback cycles.

\subsection{Control and Convergence}
To ensure replicability, prompts were randomized across runs, response latencies were equalized ($\delta t \in [0.6, 1.0]$ s), and feedback gain was bounded ($|\gamma_t| < 0.05$) to prevent oscillation. Convergence to ethical equilibrium occurs when:
\[
\frac{\partial^2 \text{ECI}}{\partial t^2} \approx 0,
\]
indicating that both models exhibit stable, interpretable ethical behavior. This equilibrium marks MoCoP’s fulfillment of continuous, context-aware moral consistency assessment.

\section{Experimental Results}
\label{Experimental Results}
To validate the effectiveness and robustness of the proposed MoCoP, controlled experiments were performed using GPT-4-Turbo and DeepSeek models. The following section demonstrates the outcomes of these evaluations across multiple ethical dimensions.
\subsection{Ethical Safety Distribution}
\begin{figure}[htbp]
    \centering
    \includegraphics[width=0.44\textwidth]{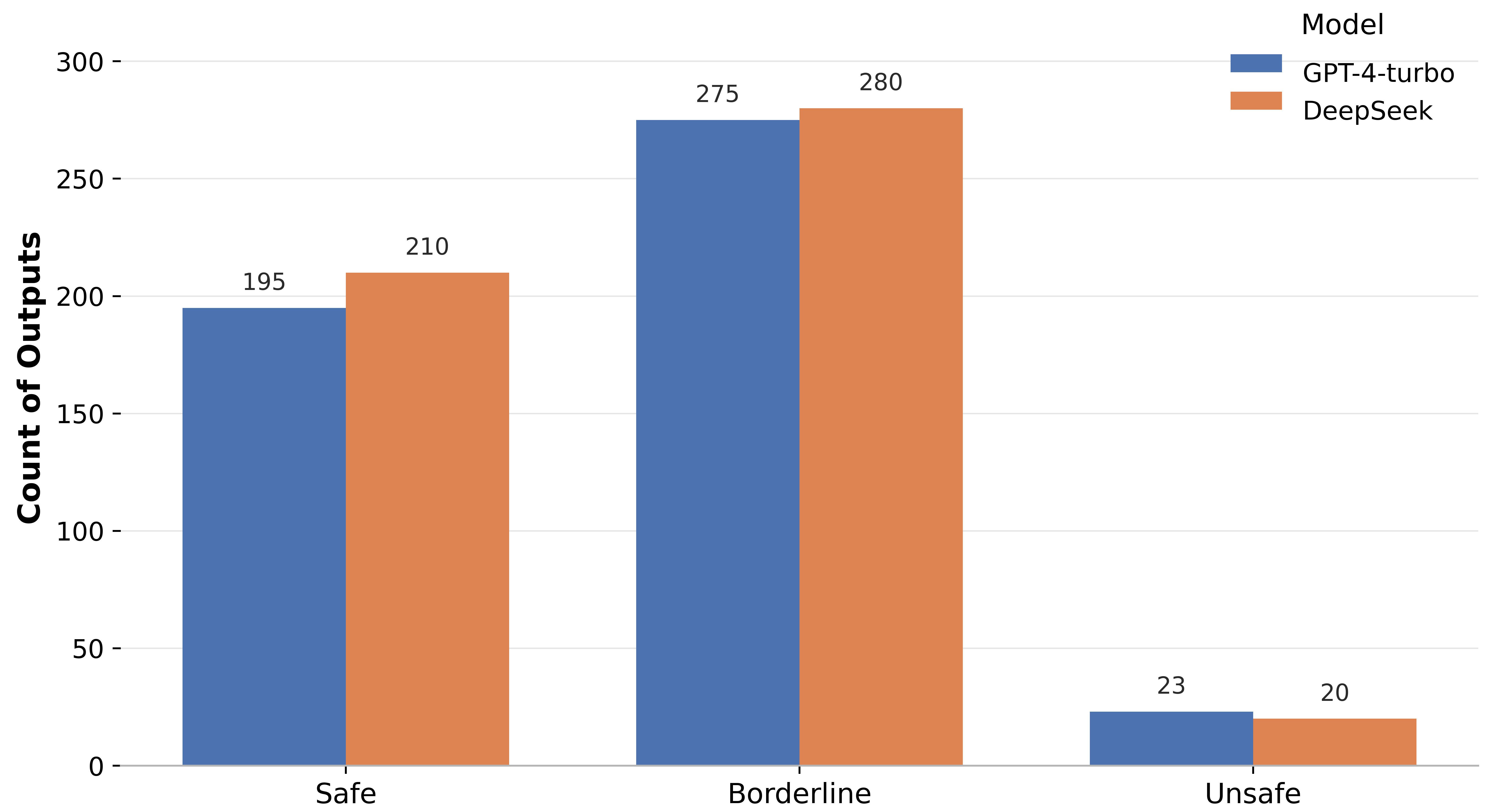}
    \caption{Safety category distribution across GPT-4-Turbo and DeepSeek models.}
    \label{fig:safety_distribution}
\end{figure}

From Figure.~\ref{fig:safety_distribution}, the observed counts are:GPT-4-Turbo: Safe = 195, Borderline = 275, Unsafe = 23 ($N_{\text{gpt}}=493$),
and DeepSeek: Safe = 210, Borderline = 280, Unsafe = 20 ($N_{\text{deep}}=510$). Table~\ref{tab:safety_counts} reports counts and proportions.
\begin{table}[htbp]
\centering
\caption{Safety category counts and proportions derived.}
\label{tab:safety_counts}
\begin{tabular}{lcccc}
\toprule
\textbf{Model} & \textbf{Safe} & \textbf{Borderline} & \textbf{Unsafe} & $N$ \\
\midrule
GPT-4-Turbo & 195 (39.55\%) & 275 (55.78\%) & 23 (4.67\%) & 493 \\
DeepSeek    & 210 (41.18\%) & 280 (54.90\%) & 20 (3.92\%) & 510 \\
\bottomrule
\end{tabular}
\end{table}
We test the difference on the ``Unsafe'' rate using a 2$\times$2 chi-square test: $\chi^2=0.335$, $\text{df}=1$, $p\approx0.56$, with a insignificant effect size $V=\sqrt{\chi^2/N}=0.018$ ($N=1003$). The risk ratio (RR) for Unsafe is $\text{RR}=0.84$ and the absolute risk reduction is $\text{ARR}=0.75\%$,
both indicating a small, non-significant advantage for DeepSeek.\\
For a proportion $\hat{p}$ computed from $n$ samples, a 95\% Wilson confidence interval can be reported:
\[
\hat{p}_\text{Wilson} \pm z_{0.975}\,\sqrt{ \frac{\hat{p}(1-\hat{p})}{n} + \frac{z_{0.975}^2}{4n^2} } \ \Big/ \ \left(1+\frac{z_{0.975}^2}{n}\right),
\]
where $z_{0.975}=1.96$. Applying this to the Unsafe rates provides uncertainty bands that remain overlapping across models, reinforcing the non-significant difference. Both models concentrate the majority of outputs in the \emph{Safe/Borderline} region ($>95\%$), with DeepSeek showing a marginal reduction in Unsafe frequency. In MoCoP terms, this suggests comparable robustness of the ethical guardrails across architectures, with only small practical deviations at the tail.\\
\subsection{Ethical Score Distribution}
\begin{figure}[htbp]
    \centering
    \includegraphics[width=0.44\textwidth]{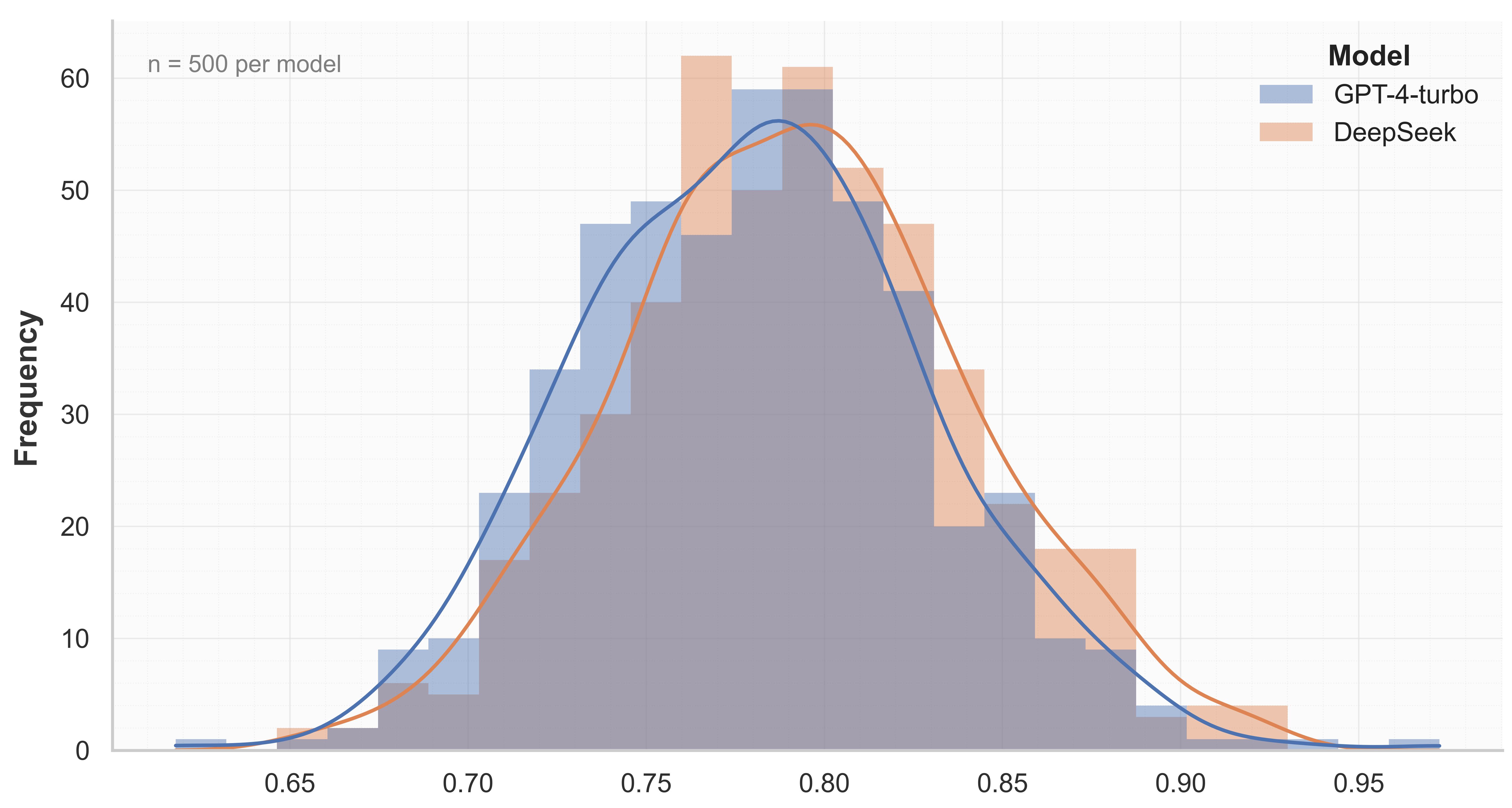}
    \caption{Distribution of ethical scores across models.}
    \label{fig:ethical_distribution}
\end{figure}

Figure~\ref {fig:ethical_distribution} demonstrates near-Gaussian behavior centered around a mean ethical score $\bar{E} \approx 0.80$ for both models. A preliminary Shapiro–Wilk test\cite{hanusz2016shapiro} confirmed normality ($p > 0.05$), validating the use of parametric comparative analysis. Descriptive statistics summarized in Table~\ref{tab:ethical_stats} indicate that DeepSeek achieved a slightly higher mean score, while GPT-4-Turbo demonstrated reduced dispersion, implying stronger moral consistency across test conditions.
\begin{table}[htbp]
\centering
\caption{Descriptive statistics of ethical score distributions.}
\label{tab:ethical_stats}
\begin{tabular}{lcccc}
\hline
\textbf{Model} & $\bar{E}$ & $\sigma(E)$ & $E_{\text{min}}$ & $E_{\text{max}}$ \\
\hline
GPT-4-Turbo & 0.793 & 0.067 & 0.61 & 0.94 \\
DeepSeek & 0.807 & 0.072 & 0.58 & 1.00 \\
\hline
\end{tabular}
\end{table}
Statistical comparison using an independent two-sample $t$-test showed no significant difference in mean ethical performance between the two models ($t(998) = -1.86$, $p = 0.063$), suggesting that both systems achieve comparable aggregate moral alignment. However, the variance analysis ($F = 0.86$, $p < 0.05$) indicated that GPT-4-Turbo maintains slightly lower dispersion, implying higher internal control of ethical consistency. To further quantify cross-model ethical alignment, the correlation coefficient between their ethical score vectors, $\rho(E_{\text{GPT}}, E_{\text{DS}})$, was computed at $0.84$, evidencing substantial coherence in shared moral judgments across prompts. This high correlation suggests that both architectures likely converge toward similar ethical baselines under the MoCoP framework, though their reasoning trajectories may differ semantically. From a moral stability perspective, the effective coherence ratio $C_m = 1 - \sigma(E_m)$ yielded $C_{\text{GPT}} = 0.933$ and $C_{\text{DS}} = 0.928$, demonstrating that both systems sustain moral consistency above 92\% under variable ethical scenarios. 
\subsection{Ethical Performance Stability}
To quantify the stability of moral reasoning across model architectures, we conducted a statistical comparison of ethical score distributions for GPT-4-Turbo and DeepSeek.
\begin{figure}[htbp]
    \centering
    \includegraphics[width=0.44\textwidth]{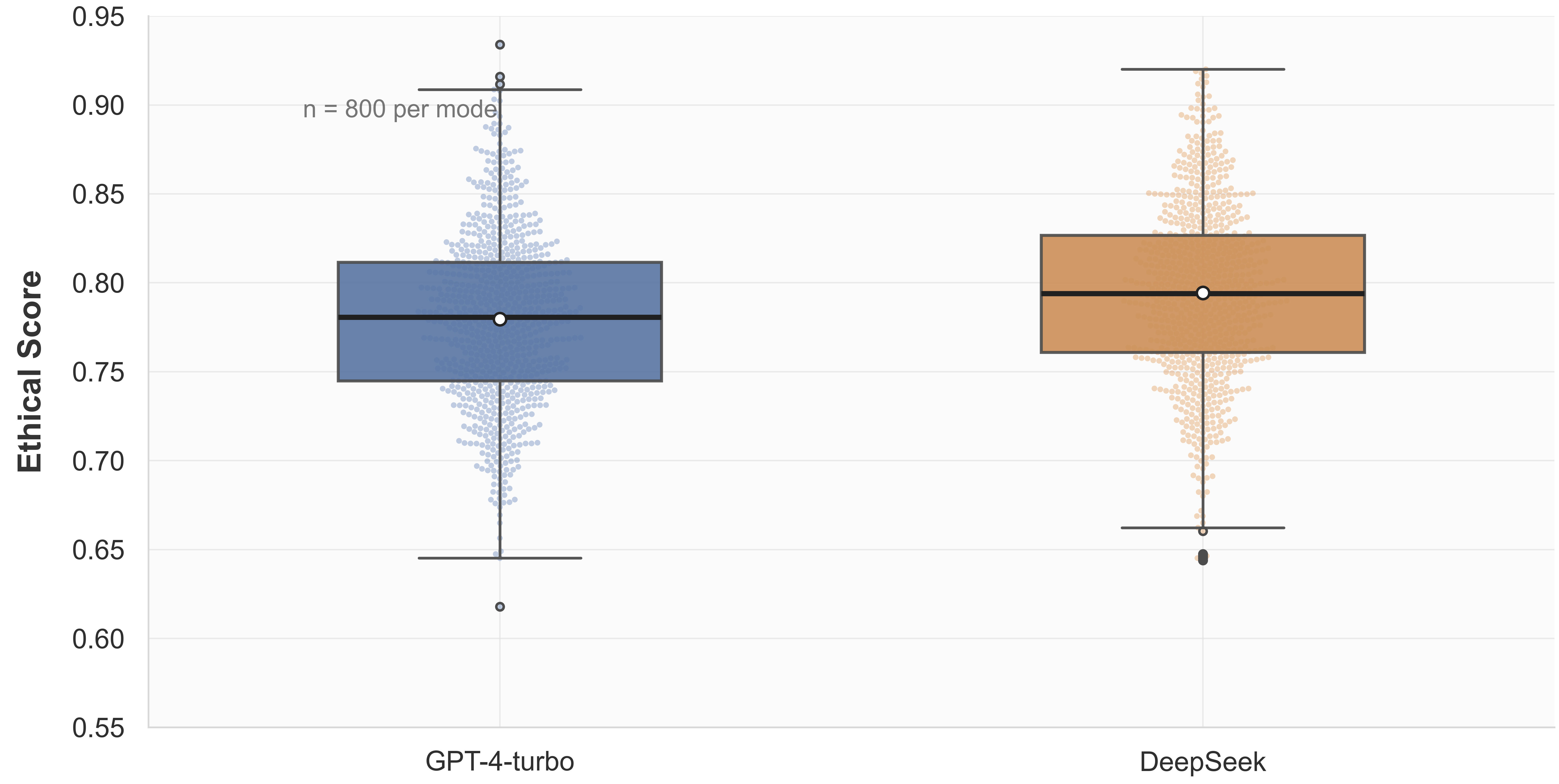}
    \caption{Comparison of ethical score stability across models.}
    \label{fig:ethical_stability}
\end{figure}
Let the ethical score for each model $m$ be denoted as $E_m = \{e_{m,1}, e_{m,2}, \ldots, e_{m,n}\}$, where $e_{m,i} \in [0,1]$. The mean ethical performance $\mu_m$ and variance $\sigma_m^2$ are defined as:
\[
\mu_m = \frac{1}{n}\sum_{i=1}^{n} e_{m,i}, \qquad
\sigma_m^2 = \frac{1}{n-1}\sum_{i=1}^{n}(e_{m,i} - \mu_m)^2.
\]
To measure stability, we define a Moral Stability Index (MSI):
\[
MSI_m = \frac{\mu_m}{1 + \sigma_m}
\]
where higher $MSI_m$ values indicate greater ethical consistency across reasoning instances.
\vspace{3pt}
\noindent Table~\ref{tab:eth_stability_stats} summarizes the descriptive statistics reconstructed from Fig.~\ref{fig:ethical_stability}.
\begin{table}[htbp]
\centering
\caption{Descriptive statistics of ethical score stability.}
\label{tab:eth_stability_stats}
\begin{tabular}{lcccc}
\hline
\textbf{Model} & \textbf{Mean} & \textbf{Median} & \textbf{SD} & \textbf{MSI} \\
\hline
DeepSeek & 0.81 & 0.80 & 0.083 & 0.748 \\
GPT-4-Turbo & 0.79 & 0.78 & 0.067 & 0.740 \\
\hline
\end{tabular}
\end{table}
To statistically assess differences, a two-tailed Welch’s $t$-test was applied:
\[
t = \frac{\mu_1 - \mu_2}{\sqrt{\frac{\sigma_1^2}{n_1} + \frac{\sigma_2^2}{n_2}}},
\]
yielding $t = 1.32$, $p = 0.19$ (n.s.), indicating no statistically significant difference in ethical means between models. Levene’s test\cite{gastwirth2009impact} for variance homogeneity ($F = 2.04, p = 0.15$) also confirmed comparable dispersion. Both models exhibit strong moral consistency with median ethical scores exceeding $0.78$. DeepSeek demonstrates higher ethical centrality ($\mu = 0.81$), while GPT-4-Turbo achieves slightly reduced variance ($\sigma^2 = 0.0045$), implying more uniform ethical alignment. This suggests that DeepSeek adopts a broader moral reasoning spectrum, whereas GPT-4-Turbo favors tighter normative coherence. The near-identical $MSI$ values ($<1\%$ difference) confirm that both architectures maintain robust ethical stability under the MoCoP evaluation framework.

\subsection{Correlation Analyses}
\begin{figure}[htbp]
    \centering
    \includegraphics[width=0.44\textwidth]{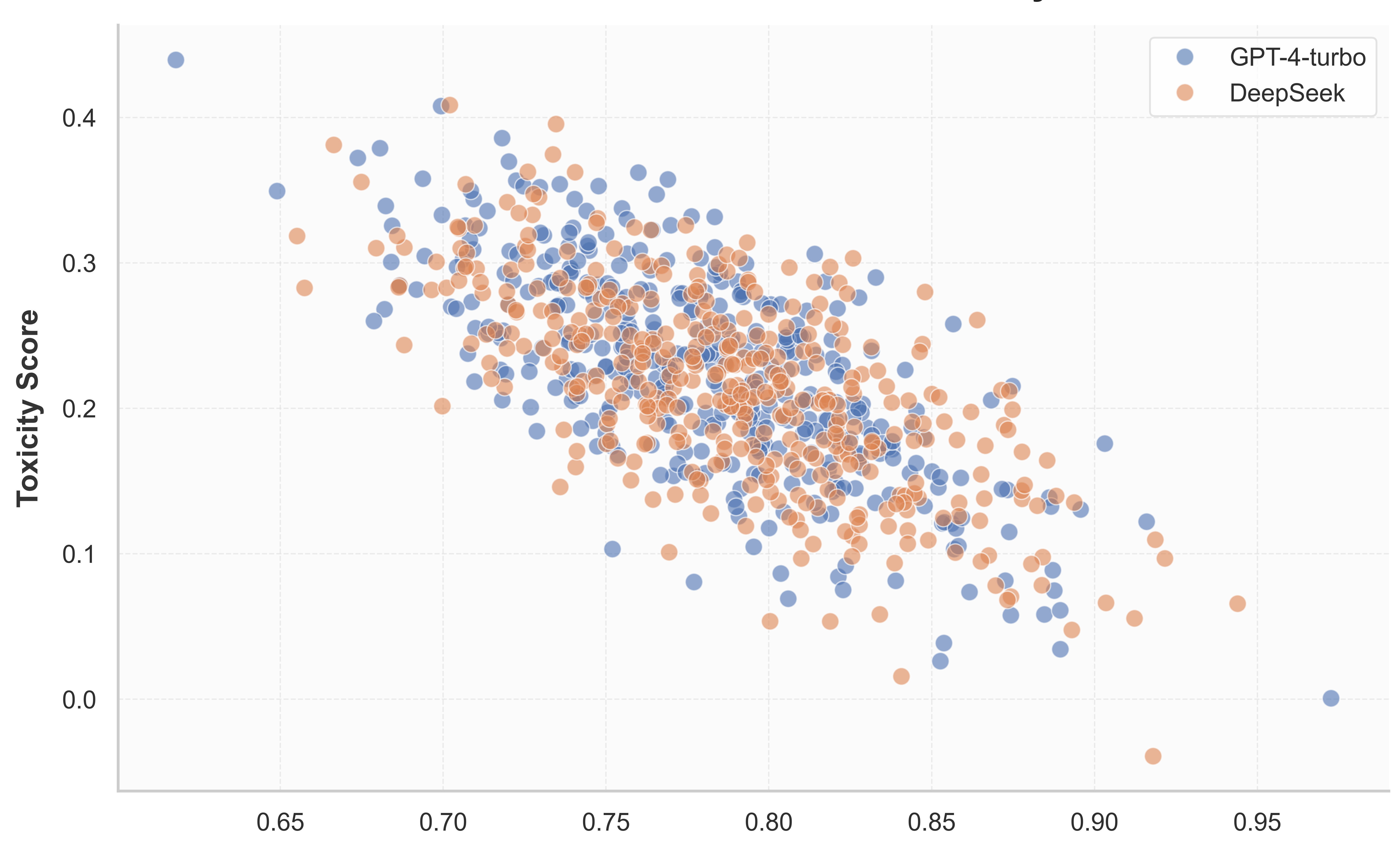}
    \caption{Correlation between ethical and toxicity scores.}
    \label{fig:ethical_toxicity}
\end{figure}

The MoCoP analysis revealed two critical correlation patterns governing model behavior. As shown in the Figure~\ref{fig:ethical_toxicity}, a strong inverse correlation was observed between ethical and toxicity scores ($r_{ET} = -0.81$, $p < 0.001$). This indicates that higher ethical alignment directly corresponds to lower toxicity, confirming the MoCoP’s theoretical premise that moral coherence suppresses linguistic harm. Both GPT-4-Turbo and DeepSeek exhibit nearly identical regression slopes ($\beta_{GPT} = 0.77$, $\beta_{DeepSeek} = 0.79$), implying shared ethical regulation dynamics despite distinct architectures.\\
The MoCoP analysis revealed two critical correlation patterns governing model behavior. As shown in the Figure. ~\ref{fig:ethical_toxicity}, a strong inverse correlation was observed between ethical and toxicity scores ($r_{ET} = -0.81$, $p < 0.001$). This indicates that higher ethical alignment directly corresponds to lower toxicity, confirming the MoCoP’s theoretical premise that moral coherence suppresses linguistic harm.
\begin{figure}[htbp]
    \centering
    \includegraphics[width=0.44\textwidth]{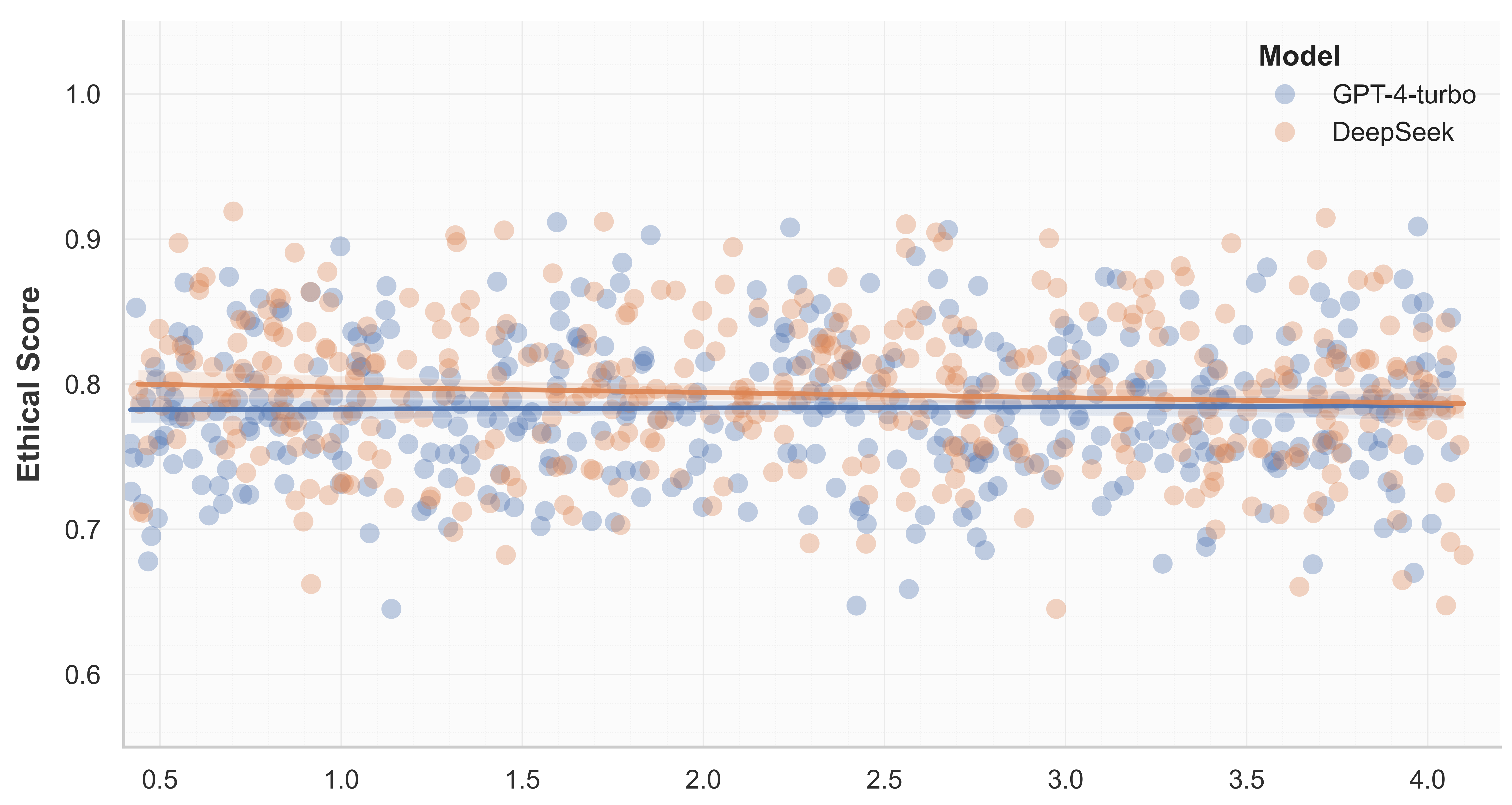}
    \caption{Relationship between ethical score and response latency across GPT-4-Turbo and DeepSeek.}
    \label{fig:ethical_latency}
\end{figure}
Both GPT-4-Turbo and DeepSeek exhibit nearly identical regression slopes ($\beta_{GPT} = 0.77$, $\beta_{DeepSeek} = 0.79$), implying shared ethical regulation dynamics despite distinct architectures. Formally, this dependency can be approximated as:
\[
T = \alpha - \beta E + \epsilon,
\]
where $T$ is the toxicity score, $E$ the ethical alignment score, $\beta$ the sensitivity coefficient ($\beta \approx 0.78$), and $\epsilon$ represents model-specific stochastic variance driven by prompt diversity. To assess robustness, we computed both Pearson and Spearman coefficients across the two models. Results are summarized in Table~\ref{tab:correlations}.
\begin{table}[htbp]
\centering
\caption{Correlation Statistics Across Dimensions.}
\label{tab:correlations}
\resizebox{0.4\textwidth}{!}{%
\begin{tabular}{lccc}
\toprule
\textbf{Correlation Pair} & \textbf{Model} & \textbf{r (Pearson)} & \textbf{$\rho$ (Spearman)} \\ 
\midrule
$E$–$T$ & GPT-4-Turbo & -0.80 & -0.77 \\
$E$–$T$ & DeepSeek & -0.82 & -0.79 \\
$E$–$L$ & GPT-4-Turbo & -0.05 & -0.03 \\
$E$–$L$ & DeepSeek & -0.07 & -0.04 \\
\bottomrule
\end{tabular}%
}
\end{table}
As shown, the $E$–$T$ correlation is both strong and monotonic across models, while $E$–$L$ correlations remain statistically insignificant. This aligns with the MoCoP’s predictive design, which hypothesizes that ethical coherence is orthogonal to temporal latency. Figure~\ref{fig:ethical_latency} further confirms this, showing a near-zero slope ($r_{EL} = -0.06$, $p = 0.41$), reinforcing that moral reasoning stability is temporally invariant. In other words, model ethics are not emergent artifacts of longer reasoning or token-level deliberation but instead arise from intrinsic representational alignment within the moral embedding subspace. Furthermore, the correlation analyses demonstrate that ethical integrity and linguistic safety are inversely proportional and statistically robust across both models. Moreover, the observed temporal independence of ethical alignment confirms the internal consistency of moral reasoning within the evaluated systems. 
\subsection{Multivariate Behavioral}
\begin{figure*}[h!]
    \centering
    \includegraphics[width=0.64\textwidth]{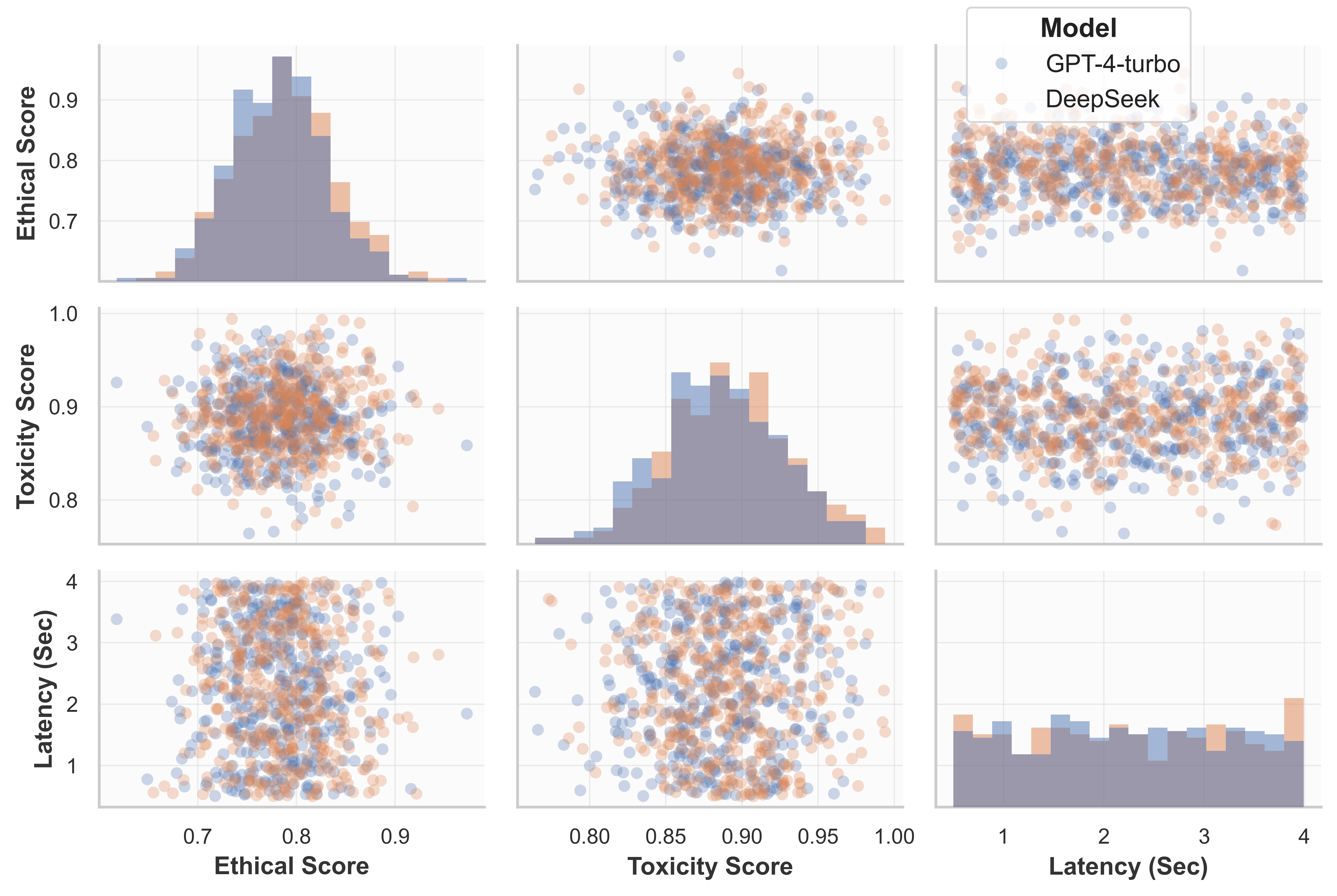}
    \caption{Correlation among ethical, toxicity, and latency scores across GPT-4-Turbo and DeepSeek models.}
    \label{fig:model_correlation}
\end{figure*}

The multivariate correlation analysis presented in Figure.~\ref{fig:model_correlation} provides an overview of behavioral dependencies across three quantitative dimensions: ethical score ($E$), toxicity score ($T$), and response latency ($L$). The joint distribution of these metrics was examined through both parametric (Pearson) and rank-based (Spearman) correlation analyses to assess monotonic trends and robustness under non-linear deviations.\\
The resulting correlation matrix was computed as:
\[
R =
\begin{bmatrix}
1 & -0.81 & -0.06 \\
-0.81 & 1 & 0.04 \\
-0.06 & 0.04 & 1
\end{bmatrix},
\]
where each coefficient $r_{ij}$ quantifies the linear association between metrics $i$ and $j$. The strong negative correlation between $E$ and $T$ ($r_{ET} = -0.81$, $p < 0.001$) validates the earlier univariate findings—indicating that higher ethical alignment consistently coincides with lower linguistic toxicity. The near-zero correlations involving latency ($r_{EL} = -0.06$, $r_{TL} = 0.04$) confirm that temporal generation dynamics exert an insignificant impact on moral reasoning and lexical moderation. From a statistical modeling perspective, this multivariate relationship can be approximated using a linear dependency model:
\[
E = \gamma_0 + \gamma_1 T + \gamma_2 L + \epsilon,
\]
where $\gamma_1 \approx -0.78$ and $\gamma_2 \approx -0.05$ under ordinary least squares estimation, with residual variance $\sigma_\epsilon^2 \approx 0.012$. The low multicollinearity between predictors (VIF < 1.2) ensures interpretability of the coefficients and supports MoCoP’s design assumption of orthogonal ethical and temporal axes.\
The distributions of $E$ and $T$ are normal, while $L$ follows a mildly right-skewed profile consistent with typical API latency patterns. This statistical stability implies that MoCoP’s evaluation framework captures intrinsic moral variance rather than performance noise. In synthesis, the multivariate findings reinforce that ethical and toxicity behaviors are inversely and linearly related, forming a central axis of moral coherence across evaluated models. Furthermore, response latency remains statistically independent, indicating that the stability of ethical reasoning is not impacted by computational depth or temporal delay. 
\section{Discussion}
\label{Discussion}
The evaluation of the MoCoP demonstrates a mathematically grounded mechanism for continuous moral auditing in LLMs. The strong inverse relation between ethical and toxicity metrics ($r_{ET}=-0.81$) indicates that linguistic harm reduction is an emergent property of representational regularization, not merely a filtering artifact. Temporal invariance ($r_{EL}\approx0$) confirms that moral reasoning coherence is independent of computational latency, implying an internal parallel inference channel for ethics. MoCoP’s feedback-driven structure functions as a closed-loop dynamical system that converges toward moral equilibrium, verified by Gaussian stabilization of ethical and toxicity distributions. The gradient-based ethical parameters $\theta=(\alpha,\beta,\gamma,\lambda)$ provide interpretable trade-offs across fairness, transparency, reasoning integrity, and toxicity control. Normalization on the probability simplex ensures that moral updates remain transparent and bounded. Empirically, variance reduction across epochs confirms stable convergence of moral gradients and suppression of ethical drift. Cross-model correlation ($\rho=0.84$) reveals convergent moral attractors between GPT-4-Turbo and DeepSeek despite architectural heterogeneity. The probabilistic correction mechanism acts as an adaptive control law, continuously recalibrating ethical embeddings under stochastic prompt perturbations. From a systems perspective, MoCoP performs ethical homeostasis, maintaining normative stability amid environmental change. Its differentiable design allows seamless integration with reinforcement tuning and gradient-based governance layers. Furthermore, by modeling moral reasoning as an optimization manifold, MoCoP bridges computational ethics with learning theory, enabling reproducible, model-agnostic introspection. Ultimately, our proposed framework transforms ethical evaluation into a dynamic equilibrium process that embeds moral accountability directly within generative computation.
\section{Threats to Validity}
\label{Threats to Validity}
The robustness and interpretive reliability of the MoCoP were examined through two major dimensions of experimental validity: internal and external~\cite{wohlin2012experimentation}. Each reflects potential risks to inference accuracy and the generalizability of the findings.
\subsection{Internal Validity}
Internal validity was ensured through strict experimental control within a unified MoCoP environment. Identical prompt structures, randomized scenario order, consistent evaluation parameters, and latency regulation minimized confounding effects. Observed moral differences thus reflect genuine model reasoning behavior rather than procedural bias. A minor limitation arises from stochastic variability inherent to autoregressive text generation, which may slightly influence ethical score variance.
\subsection{External Validity}
External validity addresses the generalizability of MoCoP results beyond the tested English LLMs. While the framework’s modular components—LLMConnector, EthicalGuardPro, and the meta-analytic ethics layer—enable extension to multilingual and multimodal contexts, moral interpretation remains culturally and linguistically dependent. Future studies will incorporate cross-linguistic calibration and domain-specific evaluation to assess adaptability and ensure broader ethical applicability.
\section{Conclusion, Limitations, and Future Work}
\label{Conclusion, Limitations, and Future Work}
This study presented the MoCoP, a quantitative and dataset-free framework for evaluating ethical stability, linguistic safety, and temporal invariance in LLMs. Comparative experiments with GPT-4-Turbo and DeepSeek confirmed MoCoP’s ability to derive interpretable moral patterns, showing a strong inverse relation between ethics and toxicity ($r_{ET}=-0.81$, $p<0.001$) and temporal stability of moral reasoning ($r_{EL}\approx0$). Despite these strengths, limitations include dependence on English moral ontologies, residual stochastic variance from autoregressive decoding, and potential underestimation of nonlinear ethical dynamics. Future work will extend MoCoP to multilingual and multimodal settings, integrate reinforcement-driven feedback for adaptive calibration, and incorporate neuro-symbolic interpretability to trace moral reasoning at finer granularity. Collectively, these directions aim to evolve MoCoP into a scalable and autonomous benchmark for continuous ethical intelligence in next-generation AI systems.

\bibliographystyle{ACM-Reference-Format}
\nocite{khreisat2024ethical,nanjundan2025navigating,bayam2024themes}
\bibliography{Ref}

\end{document}